\documentclass{new_tlp}
\makeatletter
\let\O@argtabularcr\@argtabularcr
\def\O@xtabularcr{\@ifnextchar[\O@argtabularcr{\ifnum 0=`{\fi}\cr}}
\let\O@tabacol\@tabacol
\let\O@tabclassiv\@tabclassiv
\let\O@tabclassz\@tabclassz
\let\O@tabarray\@tabarray
\def\author@tabular{\authorsize\def\@halignto{}\@authortable}
\let\endauthor@tabular=\endtabular
\def\author@tabcrone{{\ifnum0=`}\fi\O@xtabularcr\affilsize\itshape
 \let\\=\author@tabcrtwo\ignorespaces}
\def\author@tabcrtwo{{\ifnum0=`}\fi\O@xtabularcr[-3\p@]\affilsize\itshape
 \let\\=\author@tabcrtwo\ignorespaces}
\def\@authortable{\leavevmode \hbox \bgroup $\let\@acol\O@tabacol
 \let\@classz\O@tabclassz \let\@classiv\O@tabclassiv
 \let\\=\author@tabcrone \ignorespaces \O@tabarray}
\makeatother

\let\proof\relax

\usepackage{amsthm}
\theoremstyle{definition}
\newtheorem{exmp}{Example}
\usepackage{underscore}
\usepackage{graphicx}
\usepackage[inline]{enumitem}
\usepackage{hyperref}
\usepackage[table,xcdraw]{xcolor}
\usepackage{caption}
\begin{document}

  \title[ASP-based Automated Question Answering]
        {An ASP-based Approach to Answering Natural Language Questions for Texts}

  \author[D. Pendharkar et. al.]
         {Dhruva Pendharkar, Kinjal Basu, Farhad Shakerin, and Gopal Gupta \\
         The University of Texas at Dallas}




\maketitle

\begin{abstract}
    
   An approach based on answer set programming (ASP) is proposed in this paper for representing knowledge generated from natural language texts. Knowledge in a text is modeled using a Neo Davidsonian-like formalism, which is then represented as an answer set program.  Relevant commonsense knowledge is additionally imported from  resources such as WordNet and represented in ASP. The resulting knowledge-base can then be used to perform reasoning with the help of an ASP system. This approach can facilitate many natural language tasks such as automated question answering, text summarization, and automated question generation. ASP-based representation of techniques such as default reasoning, hierarchical knowledge organization, preferences over defaults, etc., are used to model commonsense reasoning methods required to accomplish these tasks. In this paper we describe the CASPR system that we have developed to automate the task of answering natural language questions given English text. CASPR can be regarded as a system that answers questions by ``understanding'' the text and has been tested on the SQuAD data set, with promising results.
    
  \end{abstract}

  \begin{keywords}
    ASP, Commonsense Reasoning, NLP, KR, Natural Language Question Answering
  \end{keywords}


\section{Introduction}

The goal of AI is to build systems that can exhibit human-like intelligent behavior. Decision-making and the ability to reason are important attributes of intelligent behavior. Hence, AI systems must be capable of performing automated reasoning as well as responding to changing environment (for example, changing knowledge). To exhibit such a behavior, an AI system needs to understand its environment as well as interact with it to achieve certain goals.
Classical logic based approaches have traditionally been used to build automated reasoning systems but have not led to systems that can be called truly intelligent. Humans, arguably, do not use classical logic in their day to day reasoning tasks. They considerably simplify their burden of reasoning by using techniques such as defaults, exceptions, and preference patterns. Also, humans use non-monotonic reasoning and can deal with incomplete information \cite{gelfond2014knowledge,how-we-reason}. All these features need to be built into an AI system, if we want to simulate human-like intelligence. It has been shown that commonsense reasoning can be realized via a combination of (stable model semantics-based) negation as failure and classical negation \cite{baral2003knowledge,gelfond2014knowledge} in ASP. ASP is a well-developed paradigm and has been applied to solving problem in planning, constraint satisfaction and optimization. There are comprehensive, well known implementations of ASP such as CLASP \cite{gebser2010gringo} and DLV \cite{alviano2011disjunctive}. Scalable implementations of ASP that support predicates (i.e., do not require grounding) and are query-driven, such as s(ASP) and s(CASP), have also been developed \cite{marple2017computing,arias2018constraint}. It is proven that ASP is well suited for representing knowledge and for modeling commonsense reasoning. 

In this paper we propose a system called CASPR\footnote{The code-base is available at - https://github.com/DhruvaPendharkar/thesis-project} (Commons-sense ASP Reasoning) to automatically convert textual knowledge into ASP programs, and use it to answer natural language questions coded as ASP queries. The problem of converting natural language text into ASP is challenging enough, however, even if we succeed in this translation task, the resulting knowledge is not enough to answer questions to the level that a human can. When we humans read a passage, we automatically draw upon a large amount of commonsense knowledge that we have acquired over the course of years in understanding the passage and in answering questions related to the passage. An automated QA system ought to do the same. CASPR, thus, resorts to resources such as WordNet \cite{miller1995wordnet}, that encapsulate some of the commonsense knowledge, to augment the knowledge derived from the text. CASPR also allows users to add commonsense knowledge---coded in ASP---manually as well.

CASPR uses default theories with exceptions and preferences to represent the textual knowledge as well as commonsense knowledge from external sources (e.g., WordNet). Also, the word-sense-disambiguation (WSD) mechanism used in CASPR is based on default theories (discussed elaborately in section 5). As we know, a default theory can be used to reason even in the absence of information and to do so negation-as-failure (NAF) is indispensable. Additionally, commonsense knowledge that we may need for textual question answering may have to represent multiple possible worlds, which cannot be elegantly modeled in traditional logic programming languages such as Prolog. An example scenario is where we may have to reason whether an animal cartoon character can talk like a human. In such a case, we will have to represent the commonsense knowledge that animals can talk in the cartoon world but not in the real world. Such reasoning over multiple worlds is most elegantly modeled in ASP. The above considerations makes answer set programming the crucial building block for the CASPR system unlike other logic programming languages (e.g., Prolog), where the support for NAF is limited or absent.

CASPR runs on the s(ASP) answer set programming system. The s(ASP) system \cite{marple2017computing} is a query-driven predicate ASP system that is scalable, in that it can run answer set programs containing predicates with arbitrary terms. Since the s(ASP) system is query-driven, it does not require grounding, an important feature needed for building large-scale natural language-based KR applications using ASP. 
The proof of the query serves as a justification, allowing us to give the reasoning behind an answer to a question in CASPR. Just like human question answering mechanism, where we are able to give a logical explanation for each answer to portray our understanding of the matter. Similarly, CASPR showcases its \textit{``true"} semantic understanding about the passage by providing justification for each answer. Also, due to its understanding CASPR is capable of answering related questions about the passage and that is a very important characteristic of any NLU application, such as a ChatBot or a SocialBot.

Our research makes several contributions: (i) it shows that with the help of novel, query-driven systems such as s(ASP), it is possible to build practical NLP applications that rely on true semantic ``understanding''; and, (ii) traditional problems of Natural Language Understanding such as word sense disambiguation can be solved quite elegantly with ASP; (iii)  finally, it demonstrates that a general (i.e., not domain-specific), scalable QA system can be built without training.

\section{Background}
We next describe some of the key technologies we employ. Our effort is based on answer set programming (ASP) technology \cite{gelfond2014knowledge}, specifically, its goal-directed implementation in the s(ASP) system. ASP supports nonmonotonic reasoning through negation as failure that is crucial for modeling commonsense reasoning (via defaults, exceptions and preferences) \cite{gelfond2014knowledge}. We assume that the reader is familiar with the basic notations of natural language processing (NLP). Expositions of NLP can be found elsewhere \cite{nlpbook}.


\noindent\textbf{Answer Set Programming (ASP): }
    An answer set program is a collection of rules of the form -
 
 \[l_0 \leftarrow l_1, \;...\,, \;l_m, \; not \,  l_{m+1}, \;...\,, \; not\, l_n. \]

    \noindent where each  \( l_i\) is a literal \cite{gelfond2014knowledge}. In an ASP rule, the left-hand side is called the \textit{head} and the right-hand side is the \textit{body}. Constraints are ASP rules without a  \textit{head}, whereas facts are rules without a \textit{body}. The variables start with an uppercase letter, while predicates and constants begin with a lowercase letter. We will follow this convention throughout the paper. The semantics of ASP is based on the stable model semantics of logic programming \cite{gelfond1988stable}. ASP supports \textit{negation as failure} \cite{gelfond2014knowledge} permitting it to elegantly model commonsense reasoning, default rules, exceptions, etc. which is responsible for CASPR's sophistication. 

\smallskip
\noindent\textbf{s(ASP) System:}
    s(ASP) \cite{marple2017computing} is a query-driven, goal-directed implementation of ASP. A query-driven implementation is indispensable for automating commonsense reasoning, 
    as traditional grounding and SAT-solver based implementations of ASP has limited scalability.
    There are three major advantages of using s(ASP): 
    \begin{enumerate*}
        \item s(ASP) does not ground the program, which makes CASPR framework fast and scalable,
        \item it only explores the parts of the knowledge base needed to answer a query, and
        \item it provides justification for an answer.
    \end{enumerate*}

\begin{figure}[ht]
    \centering
    \includegraphics[width = 0.95\textwidth]{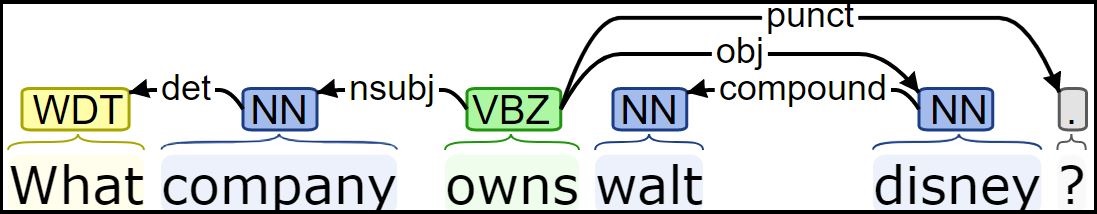}
    \caption{Example of POS tagging and dependency graph}
    \label{fig:POS}
    \end{figure}

\smallskip   
\noindent\textbf{Stanford CoreNLP Tools: }
    Stanford CoreNLP \cite{manning-EtAl:2014:P14-5}  is a set of tools for natural language processing. CASPR  uses the \textit{Parts-of-Speech (POS) tagger}, \textit{Dependency Parser}, and \textit{Named Entity Recognizer (NER)} tool from this set \cite{nlpbook}.  Based on the context, POS tagger generates the necessary parts of speech such as noun, verb, adjective, etc., for each sentence. It also identifies the question type for each question (e.g., what, where, how) and disambiguated words (e.g., \textit{block} as a \textit{verb} vs. \textit{block} as a \textit{noun}). The NER tool captures the entities present in the sentence such as location, time, etc. A dependency graph is eventually generated that shows the grammatical relations between words in the sentence. Dependency relations follow enhanced universal dependency annotation \cite{schuster2016enhanced}. Figure \ref{fig:POS} shows an example of POS tagging and dependency graph of a question.

\begin{figure}[ht]
    \centering
    \includegraphics[width = 0.95\textwidth]{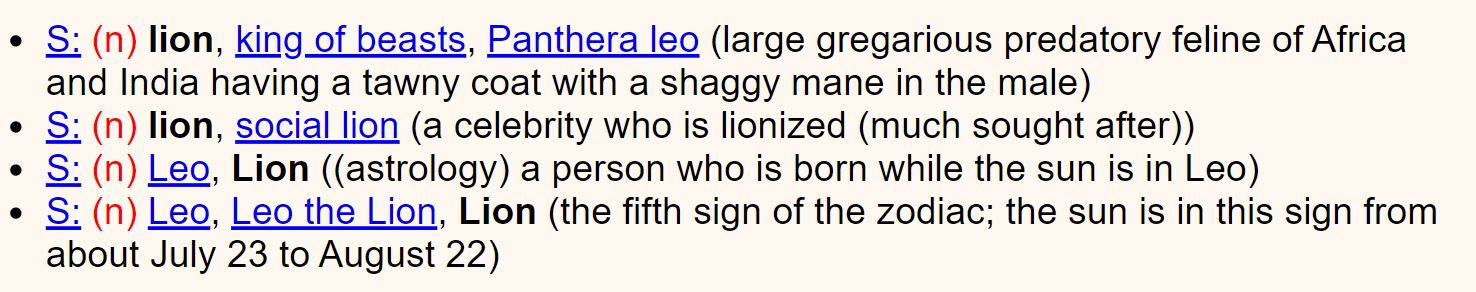}
    \caption{Entry for the concept of “lion” in WordNet}
    \label{fig:WN}
    \end{figure}

\smallskip   
\noindent\textbf{WordNet: }
 WordNet is one of the most commonly used resources in English. It is a lexical database consisting of sense-relations between English words. WordNet consists of multiple databases, one each for nouns, verbs, adjectives and adverbs. WordNet also contains a set of near-synonyms called synsets. Each database contains a set of \textit{lemmas}\footnote{\textbf{Lemmatization} is the task of determining
that two words have the same root, despite their surface differences. For example,
the word \textit{sing} is the \textbf{lemma} for words \textit{sang}, \textit{sung}, and \textit{sings} \cite{nlpbook}.}, 
each one annotated with a set of senses. A typical entry for the noun “lion” in WordNet yields the different senses shown in Figure \ref{fig:WN}. 
     
\section{CASPR: System Architecture}
\begin{figure}[ht]
\includegraphics[scale=0.45]{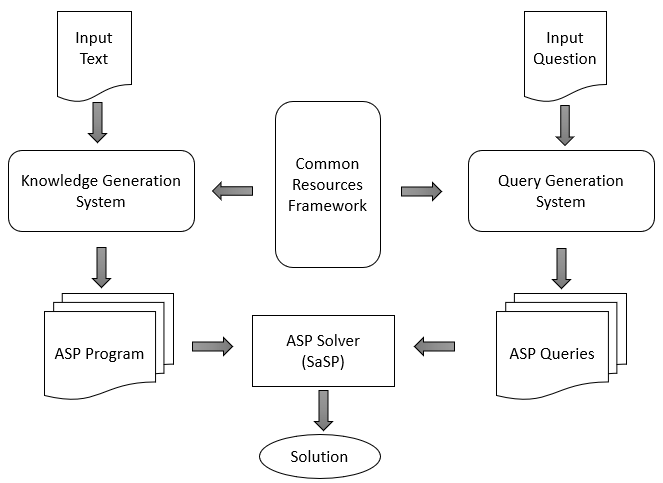}
\centering
\caption{System Architecture}
\label{fig:system}
\end{figure}

CASPR is composed of two main sub systems: the Knowledge Generation System and the Query Generation System. The architecture, as illustrated in Figure \ref{fig:system}, consists of a common resource framework shared by both these systems consisting of NLP tools such as Stanford Core NLP Tools, WordNet API as well as modules for pre-processing input text.

The Knowledge Generation System is mainly responsible for extracting knowledge from a natural language text and representing it as an answer set program. For extracting the knowledge from text and to gain more information about the input text, this component uses Stanford NLP tools like the POS Tagger, Stanford Dependency Parser, and the Stanford NER Tagger. Apart from these resources it also taps into the information that is provided by WordNet \cite{miller1995wordnet} and extracts knowledge from it, converting it into an answer set program. WordNet provides significant amount of commonsense knowledge about nouns. Unlike nouns, for verbs, at present, there are very few digital resources
available that effectively represent the information that verbs convey and that also capture relationships between them (e.g., if we \textit{drop} an object, it will \textit{fall} to the ground). Such commonsense knowledge for verbs, thus, has to be modeled manually as an answer set program, and added to the common resource framework. 

The Query Generation System automatically translates the question to an ASP query that can be executed against the ASP-coded knowledge-base generated from the textual passage augmented with ASP-coded commonsense knowledge. Execution is performed using the s(ASP) system. The solution found represents an accurate answer to the question. The query obtained from the natural language question is a conjunction of multiple sub-goals. If the query fails, then some of the sub-goals are systematically removed and the query re-executed to find less accurate answers (details are in section \ref{sec:query}). These less accurate answers are reported too, along with the level of accuracy (likely, possible, guess).

\section{CASPR: Knowledge Generation}
The Stanford Dependency Parser is used to parse the pre-processed text. A semantic graph is generated using the Stanford Typed Dependencies representation \cite{typeddependencies,typeddependencies1}.

\begin{exmp}
\textit{
``NASA carried out the Apollo program.''
}

\noindent Following is the Stanford Dependency (SD) representation:\\
\textit{nsubj(carried-2, NASA-1), root(ROOT-0, carried-2), compound:prt(carried-2, out-3),\\ det(program-6, the-4), compound(program-6, Apollo-5), dobj(carried-2, program-6)}

\end{exmp}

These dependencies map straightforwardly onto a directed graph representation in which words in the sentence are nodes in the graph and grammatical relations are edge labels. In English, most event-mentions correspond to verbs and most verbs are triggers to events. Although this is true in most cases there are other word groups that can trigger events as well. The different verbs in the sentence thus define various events that take place in the sentence and how these events are connected to each other. Consider a more complex example.

\begin{exmp}
\textit{``Miitomo, which Nintendo introduced globally in 2016, features the company's, Mii, avatar-system and lets the users communicate by exchanging personal information such as favorite movies.''}
\end{exmp}

\begin{figure}[ht]
\includegraphics[scale=0.4]{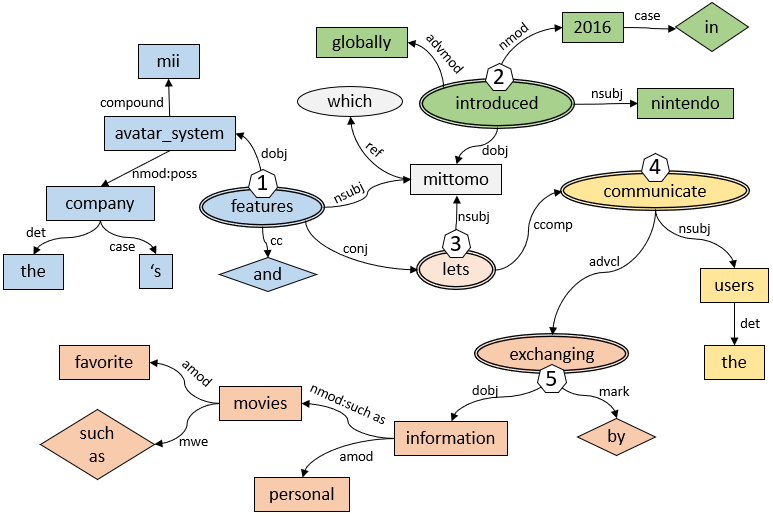}
\centering
\caption{Semantic graph representing event regions}
\label{fig:eventgraph}
\end{figure}

Figure \ref{fig:eventgraph} shows how various words in the example passage are connected to each other in the sentence. The main verbs of the sentence are ``\textit{feature}'' and ``\textit{let}'' connected by a coordinating conjunction. The verbs in Figure \ref{fig:eventgraph} represent the head of events in the sentence and are marked using event IDs. The various color regions denote the rough boundaries of these event regions. The semantic graph along with the event regions are used to create ASP facts and rules.

\subsection{Predicate Generation}
Knowledge is represented using a collection of pre-defined predicates, following the neo-Davidsonian approach \cite{neodavidsonian}. Some of the predicates capture specialized concepts such as \textit{abbreviation}, \textit{start_time} etc., whereas others are more generic like \textit{mod}, \textit{event} and so on. The generic predicates that convey information are explicitly present in the text, whereas all others model implicit information. Note that it is important to keep this predicate representation as simple as possible, so that individual pieces of knowledge can be composed using commonsense reasoning patterns. These reasoning patterns have to be kept very simple as well. Otherwise, we run the risk that we may have appropriate knowledge in our knowledge-base to answer the given question, but we fail in answering it because we are unable to compose that knowledge due to complexity of its representation. A summary of important predicates that have been used by CASPR is given below: We explain the event predicate in detail and summarize the rest (details can be found elsewhere \cite{dhruvathesis,github}).

\smallskip\noindent\textbf{Event Predicate:}
The event predicate defines an event that happens in the sentence. The verb marks the head of the event predicate. The \textit{event} predicate consists of the various actors (doers) and participants involved in the event with the signature:

\textit{event(event_id, trigger_verb, actor, participant)}

\noindent where the \textit{event_id} is an integer that uniquely identifies that event in the paragraph. The \textit{trigger_verb} denoted in the event predicate is the \textit{lemma}, i.e., the stem word of the actual word used in the sentence. The actors in the event predicate
are the subjects to the trigger\_verb in the sentence. Subjects in the sentence can be found with the help of dependencies like \textit{nsubj} and \textit{nsubj:xsubj}. Just as actors can be obtained from the subject of the sentence, the participants can be determined from direct object dependency (\textit{dobj}).

\begin{exmp}
\textit{``The American_Football_Conference's (AFC) champion team, Denver_Broncos, defeated the National_Football_Conference's (NFC) champion team, Carolina_Panthers, by 24_10 to earn AFC third Super_Bowl title.''}

\textit{event(1, defeat, denver_broncos, carolina_panthers)}

\textit{event(2, earn, afc, title)}
\end{exmp}
\noindent Also, following are the \textit{event} predicates for the passage in  \textit{Example 2} - 

\textit{event(1, feature, mittomo, avatar_system)}

\textit{event(2, introduce, nintendo, mittomo)}

\textit{event(3, let, mittomo, null)}

\textit{event(4, communicate, users, null)}

\textit{event(5, exchange, null, information)}

\noindent Further, we can get richer information about the event by generating duplicate event predicates for each modifier for the actor as well as the participants involved in the event. To generate such event predicates, we use the \textit{amod} or \textit{nummod} dependencies for the actors and the participants and create compound atoms from the modifiers and their governors. Consider the following duplicate event predicate:

\textit{event(2, earn, afc, third_super_bowl_title)}

\noindent Note that in absence of information, the default value for the actor and the participant field maybe \textit{null}. A \textit{null} value indicates that either the term is absent for the event or the system was not able to determine it.

\noindent\noindent\textbf{Property Predicate:}
The property predicate elaborates on the properties of the modified noun or verb. The modifier in this case is generally a prepositional phrase in the sentence. A property predicate is coupled with an event and describes the modification only for that event.

\begin{exmp}
\textit{``The game was played on February 7, 2016, at Levis_Stadium, in the San_Francisco_Bay_Area, at Santa_Clara in California''}

\textit{_property(2, play, on, `february_7_2016')}

\textit{_property(2, play, at, levis_stadium)}

\textit{_property(2, play, in, san_francisco_bay_area)}

\textit{_property(2, play, at, santa_clara)}

\textit{_property(2, santa_clara, in, california)}

\end{exmp}


\smallskip\noindent\textbf{Modifier Predicate:}
The modifier predicate is used to model the relationship between adjectives and the nouns they modify and between verbs and their modifying adverbs. 
\begin{exmp}
\textit{``The Amazon_rainforest, also known in English as Amazonia or the Amazon_Jungle, is a moist broadleafed forest that covers most of the Amazon_basin of South_America.''}

\textit{_mod(forest, broadleafed)}

\textit{_mod(forest, moist)}

\textit{_mod(know, also)}
\end{exmp}

\smallskip\noindent\textbf{Possessive Predicate:}
The possessive predicate is used to model the genitive case in English. It is used to show possession or a possessive relation between two entities in the sentence. 

\begin{exmp}
\textit{
``The American_Football_Conference's (AFC) champion team, Denver_Broncos, defeated the National_Football_Conference's (NFC) champion team, Carolina_Panthers, by 24_10 to earn AFC third Super_Bowl title''}

\textit{_possess(american_football_conference, team)}

\textit{_possess(national_football_conference, team)}

\textit{_possess(american_football_conference, denver_broncos)}

\textit{_possess(national_football_conference, carolina_panthers)}
\end{exmp}

\smallskip\noindent\textbf{Instance Predicate:}
The instance predicate models the concept of an instance. As an example, \textit{red} is an instance of a \textit{color}. 

\begin{exmp}
\textit{
``Nikola_Tesla was a serbian-american inventor, electrical engineer, mechanical engineer, physicist, and futurist''}

\textit{_is(nikola_tesla, inventor).}

\textit{_is(nikola_tesla, serbian_american_inventor).}
\end{exmp}
\noindent In the above example we see that the verb (\textit{is}) is associated with other concepts like engineer, physicist, and futurist using the conjunction. Thus, we can extend the definition of the instance predicate to also include these other facts:

\textit{_is(nikola_tesla,electrical_engineer),}

\textit{_is(nikola_tesla, futurist),}

....

\noindent Adding these facts makes the knowledge base richer which is now able to infer many other things about the passage. Another case, where we can generate the instance predicate is in cases where multi-word expressions like \textit{such as}, or \textit{like} are used to compare two concepts to be equivalent. Consider the following sentence as an example for the same.

\begin{exmp}
\textit{
``Miitomo, which Nintendo introduced globally in 2016, features the company's, Mii, avatar-system and lets the users communicate by exchanging personal information such as favorite movies.''
}

\textit{_is(movie, personal_information)}

\textit{_is(favorite_movie, personal_information)}

\end{exmp}

\noindent In the above example we use the expression “such as” to compare two concepts to be equivalent in the sense that one concept is a likely equivalent of the other. In such cases the generated instance predicate is shown above. Such predicates can be generated using the \textit{nmod, mwe,} and the case dependencies.

\smallskip\noindent\textbf{Relation Predicate:}
The relation predicate is used to connect two concepts in events. This predicate is generated to model mainly two relations, i.e., \textit{dependent clauses} and \textit{conjunctions}. The signature of the relation predicate can be given as follows.

\textit{_relation(independent_entity, dependent_clause_id, relation_type)}

\noindent Dependent clauses can be of two types depending upon their governors, i.e., adjective clauses or adverb clauses. Adjective clauses are dependent clauses that modify a noun, whereas adverb clauses modify a verb. The \textit{independent_entity} defined in the signature can be either the noun that is modified or the event ID of the verb that is modified. The \textit{dependent_clause_id} is always the ID of the verb that is the head of the dependent clause. The \textit{relation_type} define the relation between the dependent and the independent clause. This system recognizes 3 types of relations viz. \textit{_clause}, \textit{_clcomplement} and \textit{_conj}. All these relations are discussed below with examples.

As mentioned above the \textit{_clause} relation is generated for adjective and adverb clauses. Examples for both these clauses is given as follows.

\begin{exmp}
\textit{
``The American_Football_Conference's (AFC) champion team, Denver_Broncos, defeated the National_Football_Conference's (NFC) champion team, Carolina_Panthers, by 24_10 to earn AFC third Super_Bowl title''}

\textit{_relation(1, 2, _clause)}

\textit{event(1, defeat, denver_broncos, carolina_panthers)}

\textit{event(2, earn, afc, title)}

\end{exmp}

\noindent In the above given sentence, there are two clauses, one headed by the verb \textit{``defeat''} and the other by the verb \textit{``earn''}. Here, the clause headed by \textit{``defeat''} is the main clause whereas the adverb clause headed by \textit{``earn''} is the dependent clause. Such a relation can be found out using the dependency \textit{advmod}. Similarly, consider the adjective clause in the following sentence.

\begin{exmp}
\textit{
``The American_Broadcasting_Company (ABC), stylized in the network's logo as ABC since 1957, is an American commercial broadcast television network''}

\textit{_relation(american_broadcasting_company, 1, _clause)}

\textit{event(1, stylize, null, null)}

\end{exmp}

\noindent In this sentence, the noun \textit{``American_Broadcasting_Company''} is modified by the dependent clause headed by the verb \textit{``stylize''}. Such a predicate can be modeled using the \textit{acl} dependency relation. One of the other relations is \textit{_clcomplement}. This type of relation models both the clausal complement \textit{(comp)} as well as the open clausal complement \textit{(xcomp)}. Such a relation is used to search for the subject of the dependent clause. An example of such a relation is given as follows.

\begin{exmp}
\textit{
``The ideal thermodynamic cycle used to analyze the process is called the Rankine_Cycle''}

\textit{_relation(1, 2, _clcomplement)}

\textit{event(1, use, null, null)}

\textit{event(2, analyze, null, process)}
\end{exmp}

\noindent Conjunctions are words that connect two or more clauses. The relation predicate is also used to model this relation between any two clauses. An example of such a predicate is given below.

\begin{exmp}
\textit{
``Water heats and transforms into steam within a boiler operating at a high pressure''}

\textit{_relation(2, 3, _conj)}

\textit{event(2, heat, water, null)}

\textit{event(3, transform, water, null)}

\end{exmp}

\noindent In the above example we use the \textit{_conj} relation type to model the conjunction relation between the verbs ``heat'' and ``transform''. As seen above the \textit{_conj} relation type uses ID’s for verb conjunctions whereas actual nouns for noun conjunctions.

\smallskip\noindent\textbf{Named Entity Predicate:}
CASPR uses the Named Entity Tagger (NER) \cite{nertagger} to get information about entities in the text. The Named Entity Tagger marks various classes like LOCATION, PERSON, ORGANIZATION, MONEY, PERCENT, TIME in the text. We make use of these tags to generate facts of the form \textit{concept(instance)}. 
For an example, \textit{location(san_francisco)} is captured and represented as a fact if a sentence has the word \textit{`San Francisco'} and NER captures that it is a location name.
These facts together with the rules generated by the ontology help in reasoning about the text. 

\smallskip\noindent\textbf{Special Predicates:}
Special predicates have been used in the system to model concepts that are patterns and are understood by humans implicitly. As grammatical relations in the sentence do not convey the meaning of these concepts, they must be extracted explicitly. Some of them include abbreviations, time spans and so on. 
Also, for each special pattern, we have added either ASP facts in the system or generic default rules, so that we need to add it once and can be used for any text without any change.
The more of these patterns are learned by the system the better it can reason like humans. 

\vspace{-0.1in}

\begin{itemize}
    \item \textbf{Time Span Predicates: } Time spans are sometimes expressed in text using the bracketed notion. The meaning of these time spans depends upon the noun that they follow. If the noun is a person, then it may mean that the time span indicates the birth and the death date of the person. On the contrary if the noun is an organization or a product or a project then the time span indicates the start and end date for the organization or the project. 
    As there can be multiple other possibilities, we have taken the default ones (mostly-used) unless mentioned about the exceptions and we found from the results that this works really well.
    A time-span is generally of the format ``(DATE - DATE)'', where the first date is the start date and the second is the end date. The signature for the start and the end date can be given as:
    
    \textit{~~~~~~~_start_date(entity, date)}
    
    \textit{~~~~~~~_end_date(entity, date)}

    For an example, the time-span predicates for the sentence - \textit{``Project_Mercury was followed by the two-man Project_Gemini (1962 – 1966)''} is --- \textit{_start_date (project_gemini, 1962), ~_end_date (project_gemini, 1966)}.
     \item \textbf{Date Part Predicates: }
     The time predicates that have been detected from the Named Entity Tagger, can be utilized to get the day, month, and year parts from the time. This information about the various parts of time is understood implicitly by humans. We can use simple information extraction techniques to find out the year, month, and day predicates from the time predicate. The signature of these predicates can be given as follows: 
     
     \textit{~~~~~~~day(time, day)}
     
     \textit{~~~~~~~month(time, month)}
     
     \textit{~~~~~~~year(time, year)}
     
      For an example, the date-part predicates for the time phrase \textit{`10_november_1483'}  is - \textit{day(`10_november_1483',~10),~month(`10_november_1483',~november),~year(`10_november_1483',~1483)}.
      \item \textbf{Concept Predicates for Appositional Modifiers: }
      Appositional modifiers directly follow the noun they describe. We could use this information to model the fact that most appositional modifiers follow the \textit{instance_of} pattern. This is used to generate concept predicates from the appositional modifiers. Consider the following example for creating a concept predicate.
      \begin{exmp}
        \textit{
            ``Jim's brother, Sam, is coming to town today.''}

            \textit{~~~~brother(Sam)}

      \end{exmp}
      \noindent In the above sentences, we have a relation of an appositional modifier between the words \textit{``brother''} and \textit{``Sam''}. This tells us that these words may share an \textit{instance_of} relation between each other, which leads us to generate the above facts from the appositional modifier.

       \item \textbf{Abbreviation Predicate: }
       Sometimes, texts contain abbreviation of words. It becomes very important for a system to understand the meaning of these abbreviations for querying. This is because the queries asked can sometimes contain short forms for organizations and projects rather than their long forms. Abbreviations can be detected by either doing pattern matching i.e. looking for the pattern ``CONCEPT (ABBREVIATION)'' or by using the \textit{appos} dependency relation. The signature for the abbreviation predicate is as follows:
       
       \textit{~~~~~~~_abbreviation(short_form, long_form)}
       
       \item \textbf{Number Predicate:} The number predicate is used to identify numbers. Identifying a word to be a number helps in constraining the domain for certain queries. Other than that, for questions like \textit{``how many''} and \textit{``how much''} it is more accurate to give a number as the answer. The signature for the number predicate can be given as:
       
       \textit{~~~~~~~number(value)}
       
       For an example, the number predicates for the sentence - \textit{``Kenya covers 581,309 sq.km. and had a population of approximately 45 million people in July 2014.''} is --- \textit{number(`581,309'), number(45), number(million)}.

\end{itemize}

\subsection{Commonsense Knowledge Generation}
 
We  make use of additional knowledge sources such as the WordNet to gain supplementary information (commonsense knowledge) about the concepts in the input passage. CASPR uses the Hypernym relation from WordNet to build its ontology, coded as an answer set program. 

To generate ontology rules, we use standard knowledge patterns from answer-set programming like the preference pattern and the default reasoning pattern \cite{gelfond2014knowledge,zhuo-iclp2016}. In general a concept is represented in our work using the following signature:

\textit{concept(concept_instance, instance_sense)}

\noindent For example: \textit{lion(simba, noun_animal).}

\begin{figure}[ht]
\includegraphics[scale=0.4]{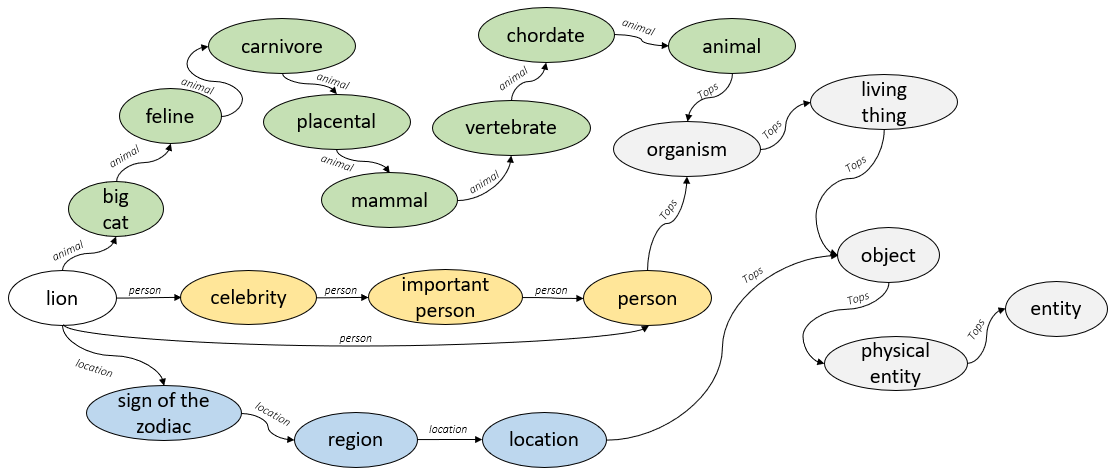}
\centering
\caption{Concept Graph of Hypernym Relation}
\label{fig:hypernymgraph}
\end{figure}

\begin{figure}[ht]
\includegraphics[scale=0.3]{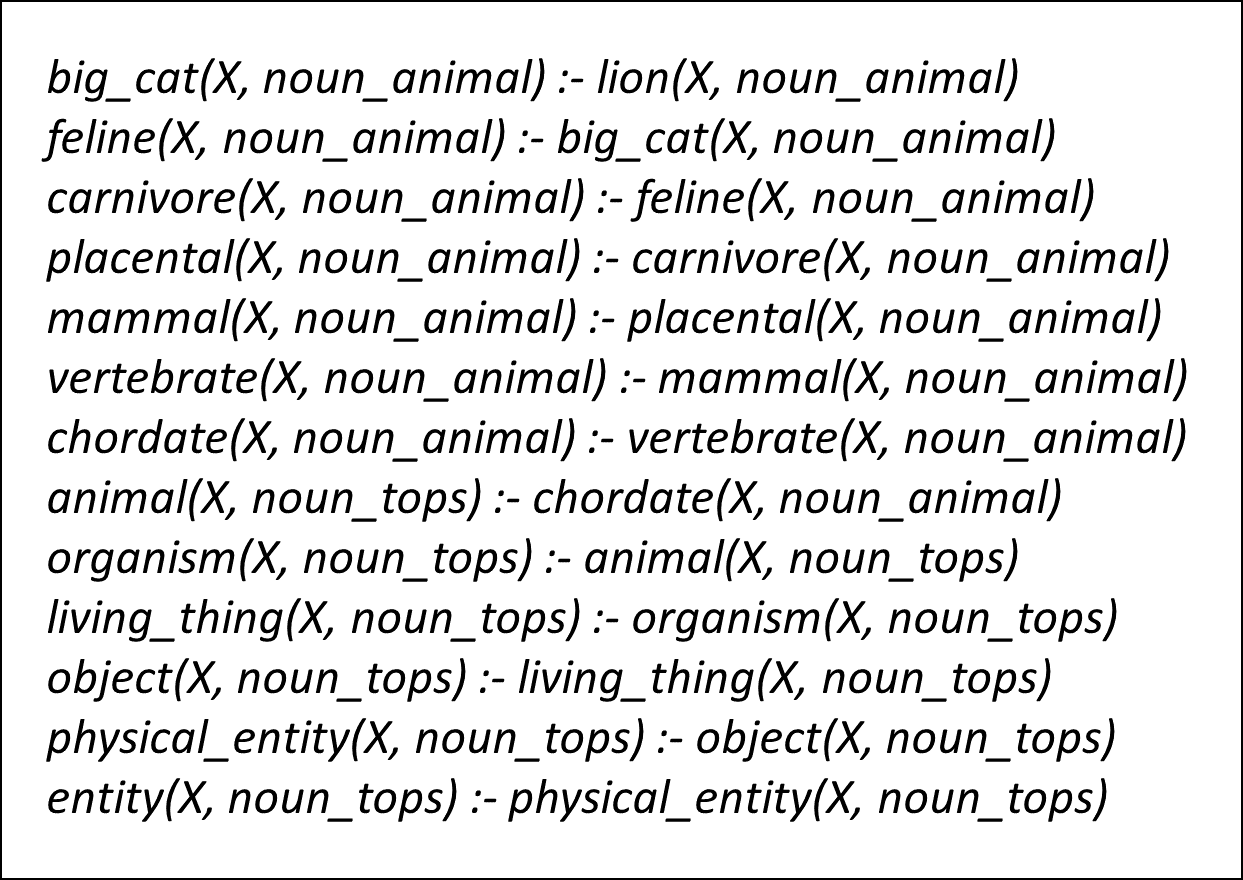}
\centering
\caption{ASP Rules representing the animal branch of ``lion"}
\label{fig:hypernymrules}
\end{figure}

Hypernyms can be used to infer various properties of and functions about concepts. Hypernyms make use of the generalization principle to transfer properties from more general concepts to their specific concepts. There are three steps required to do so (i) Identify the concepts from the passage and generate a hypernym graph; words may be used in multiple senses (e.g., the word \textit{lion} has four commonly used senses); example graph for word lion highlighting 4 different senses is shown in fig. \ref{fig:hypernymgraph}.  (ii) Aggregate the concepts into common base concepts. (iii) Generate hypernym rules. Figure \ref{fig:hypernymrules} shows ASP rules  generated by CASPR for one of the senses of the concept \textit{lion}. 
Due to the space constraint, we have omitted a few details, however more illustrations can be found in the thesis \cite{dhruvathesis}.

\medskip
\noindent\textbf{Adding Commonsense Knowledge Manually:}
Note that at the moment we are only using WordNet, however, other knowledge resources, such as YAGO \cite{yago}, VerbNet \cite{verbnet}, never-ending language learner \cite{nell}, etc., can be incorporated as well to extract additional commonsense knowledge. Doing so is relatively straightforward, as it mostly involves syntactic transformation to ASP syntax. Note that, currently, commonsense knowledge about verbs is entered manually. An example of adding knowledge manually is the following:

\noindent The term Nikola\_Tesla is similar to Tesla:
   
\textit{_similar(tesla, nikola_tesla).}

\noindent A team X can represent an organization Y, if X is possessed by Y:
   
\textit{event(E, represent, X, Y):-_possess(Y, X), organization(Y), team(X),} 

\textit{~~~~~~~~~~~~~~~~~~~~~~~~~~~~~~~~~~~~~~~~~~~~~~~~~~~~~~~~~~~~~~~~~~~~not ab_event(E, represent, X, Y). }

\noindent 
These handcrafted knowledge are represented using either ASP facts or as default rules with exceptions and also it uses generic predicate names. So, these manually entered commonsense knowledge added once can reside in the knowledge-base permanently and can be used in any other application without any change. This approach is very similar to humans, where we learn knowledge once, and may use it later in another context. 
\section{Word Sense Disambiguation}

Word sense disambiguation (WSD) is the task of selecting the best sense out of a collection of senses applicable to a concept. 
When we query a concept from WordNet lexicon, we get a list of senses for a specific concept ordered from the most used to the least used.
WSD is a very common problem in NLP applications and has many statistical  solutions that have been developed. 
We develop a method using negation as failure, classical negation and preferences that models WSD correctly. 
Essentially, contextual knowledge is used to disambiguate the word sense, in a manner that humans use. We explain it next. 

\smallskip\noindent\textbf{Representation of Word Senses:}
Word senses are represented using two different logic patterns in our work. Using both these patterns together, senses are selected for the various concepts in the text which activate their hypernym relations. 

In a natural language, a concept (i.e., word) can have multiple senses and one of those senses must have been used in the sentence based on the textual context. The task is to identify that sense. The pattern discussed in this section tries to discover the correct sense for a concept using the characteristics of the sense.
Its template can be given as follows:

\textit{c(X, $s_i$) :- c(X), characteristics($s_i$, X), not -c(X, $s_i$).}

\noindent In the above template, we are trying to prove that ``\textit{X} is an instance of concept \textit{c} with sense $s_i$''. Here every concept \textit{c} has one or more senses denoted by $s_i$. The above rule states that \textit{X}  is an instance of the concept \textit{c} with the sense $s_i$ only if we can prove that \textit{X} is some instance of concept \textit{c}, \textit{X} has all the characteristics required to be of sense $s_i$, and we cannot prove that \textit{X} is definitely \textit{not} an instance of concept c with sense $s_i$. 
Please note that the ``\textit{characteristic}'' predicate becomes true for the instance \textit{X} with sense $s_i$ iff it is supported by the knowledge represented from the passage (the context of the passage) in ASP form. 
Such rules are generated for every sense $s_i$ of the concept in the order of senses from the most used sense to the least used sense. The first term given by `\textit{c(X)}' is responsible to short circuit and fail the rule if the instance \textit{X} does not belong to the class \textit{c}. The second term, `\textit{characteristics($s_i$, X)}', is the main predicate which tries to prove that \textit{X} shows the characteristics of having sense $s_i$. It is this predicate that can be added either manually or using other sources to prove the sense. The third term of the body is a strong exception against the head of the rule, that fails the rule if an exception is found against the application of the sense.

\noindent According to WordNet, the concept `\textit{tree}' has three senses, S = \{\textit{plant, diagram, person}\}, ordered according to their frequency of use. Thus, using the above-mentioned template for the sense the three rules generated for the tree concept can be given as follows

\textit{tree(X, plant) :- tree(X), characteristics(plant, X),}

\textit{ ~~~~~~~~~~~~~not -tree(X, plant).}

\textit{tree(X, diagram) :- tree(X), characteristics(diagram, X),}

\textit{ ~~~~~~~~~~~~~not -tree(X, diagram).}

\textit{tree(X, person) :- tree(X), characteristics(person, X),}

\textit{ ~~~~~~~~~~~~~not -tree(X, person).}

\smallskip\noindent\textbf{Preference Patterns for Senses:}
We humans perform word sense disambiguation in our day-to-day life by taking cues from the context. Over a period of time we learn which senses are more common than others and develop a preference order.  Consider a concept \textit{c} having three senses \textit{s$_1$}, \textit{s$_2$} and \textit{s$_3$} ordered according to the frequency of their use from the most used to the least used. We first assume the sense to be \textit{s$_1$}, unless we know that \textit{s$_1$} is not the sense from some other source. Then we move on to the next sense \textit{s$_2$} unless we know that both \textit{s$_1$} and \textit{s$_2$} cannot be applicable. Then finally we choose \textit{s$_3$}. This process of elimination of senses and choosing senses according to preferences can be modeled as the following ASP code template.

\textit{c(X, $s_p$) :- c(X), not -c(X, $s_p$),}

\textit{~~~~~~~-c(X, $s_1$), -c(X, $s_2$), ..., -c(X, $s_{p-1}$),}

\textit{~~~~~~~not c(X, $s_{p+1}$), not c(X, $s_{p+2}$), ..., not c(X, $s_n$).}

\noindent The above skeleton is applied for all the senses of concept \textit{c} in order of preference. The above template represents the rule generated for the $p^{th}$ sense of the concept \textit{c} such that $1 < p < n$, where \textit{n} is the total number of senses of concept \textit{c}. If \textit{p = 1} then the template omits the classical negation terms as follows:

\textit{c(X, $s_1$) :- c(X), not -c(X, $s_1$), not c(X, $s_2$), }

\textit{~~~~~~~~~~~~~~~not c(X, $s_3$), ..., not c(X, $s_n$).}

\noindent Similarly, if \textit{p = n}, then the template omits the negation as failure terms, given as:

\textit{c(X, $s_n$) :- c(X), not -c(X, $s_n$), -c(X, $s_1$),}

\textit{~~~~~~~~~~~~~~~-c(X, $s_2$), ... -c(X, $s_{n-1}$).}

\noindent We can apply this pattern to the `tree' concept as an example. A tree can be a living object (plant), a diagram, or a person (Mr. Tree).

\textit{tree(X, plant) :- tree(X), not -tree(X, plant),}

\textit{ ~~~~~~~~not tree(X, diagram),}
\textit{ not tree(X, person).}

\textit{tree(X, diagram) :- tree(X),}
\textit{ not -tree(X, diagram),}

\textit{ ~~~~~~~~-tree(X, plant),}
\textit{ not tree(X, person).}

\textit{tree(X, person) :- tree(X),}
\textit{ not -tree(X, person),}

\textit{ ~~~~~~~~-tree(X, plant),}
\textit{ -tree(X, diagram).}

\noindent This pattern is responsible for assigning at least one sense for every concept in the text. This preference pattern along with the property pattern mentioned previously helps disambiguate word sense just as humans do.

\section{CASPR: Query Generation from Natural Language Question} \label{sec:query}

Once knowledge has been generated from the text and auxiliary sources, our next task is to translate the question we want to answer into an ASP query. 

Since a question is also a sentence, it is processed in the same way that any other sentence would be, as mentioned before in this paper. This means that a semantic graph is generated for every question, and event regions are created within the question assigning event ids to different parts of the question. To generate an ASP query from the semantic graph the following steps are applied: (i) Question understanding (ii) Generation of query predicates (iii) Applying base constraints (iv) Combining constraints. Currently, this module is only built to deal with simple interrogative sentences but can be extended to deal with more complex questions. 
In the following subsections, we have explained all the steps of ASP query generation process from natural language questions. For more details users are requested to refer to the thesis (chapter 7) \cite{dhruvathesis}.  

\medskip\noindent\textbf{Question Understanding:}
For analyzing questions, we obtain four kinds of information from the question, namely, the question word, the question type as mentioned in the previous section, the answer word or the focus of the question, and the answer type. 

The question word is the Wh-word found in the question. In general, Wh-words can be found out by looking at the following POS Tags on the words: WDT, WP, WP\$ and WRB. If none of the tags are found, then the questions may have a  copula \cite{nlpbook} as its question word or a modal as its question word. The question type contains a set of predefined question types that help in further processing. These include WHAT, WHERE, WHO, WHICH, WHEN, HOW\_MANY, HOW\_MUCH, HOW\_LONG, HOW\_FAR, and UNKNOWN. Once the main question word is found, we can use the word and its relations to determine the exact question type. The third type of information we try to extract is the answer word or the focus of the question. The answer word is the word in a question that tells us what kind of answer is expected from the question. Not all types of questions require an answer word, so in those cases this information is null. The fourth and the last information that we extract from the question is the answer type. The answer type depends on the question type. For most of the question types the required answer type is predefined. The answer types supported by the system are as follows SUBJECT, OBJECT, PLACE, PERSON, TIME, YEAR, DAY, MONTH, NUMBER (non-negative integers) and UNKNOWN. Table 1 shows expected answer types depending on question types. Please note that, in the table, \textit{``variable''} answer type means it can be any type and not bounded by a domain of answers. With the help of this basic analysis, we start generating the predicates for query generation.


\begin{table}[ht]
\begin{center}
    
\label{answertypetable}
\begin{tabular}{|l|l|}
\hline
\rowcolor[HTML]{EFEFEF} 
\multicolumn{1}{|c|}{\cellcolor[HTML]{EFEFEF}{\color[HTML]{000000} \textit{\textbf{Question Type}}}} & {\color[HTML]{000000} \textit{\textbf{Expected Answer Type}}} \\ \hline
WHERE                                                                                                & PLACE                                                         \\ \hline
WHO                                                                                                  & PERSON                                                        \\ \hline
WHEN                                                                                                 & TIME                                                          \\ \hline
HOW\_MANY                                                                                            & NUMBER                                                        \\ \hline
HOW\_MUCH                                                                                            & NUMBER                                                        \\ \hline
HOW\_LONG                                                                                            & NUMBER {[}length{]}                                           \\ \hline
HOW\_FAR                                                                                             & NUMBER {[}distance{]}                                         \\ \hline
WHAT                                                                                                 & ** variable **                                                \\ \hline
WHICH                                                                                                & ** variable **                                                \\ \hline
UNKNOWN                                                                                              & UNKNOWN                                                       \\ \hline
\end{tabular}
\caption{Expected Answer Types for Question Types}
\end{center}
\end{table}

\medskip\noindent\textbf{Generating Query Predicates:}
Using the information that we gathered during Question Understanding, we start generating predicate facts for all the words in the question. We will use predicates mentioned previously in the paper as a reference. There are some changes that need to be made in a few predicates, e.g., event predicate, property predicate, etc. Others (possess, mod, and named entity predicates) are generated as discussed earlier.

\smallskip\noindent\underline{\it Event Predicate:}
The event predicate in general can be given as follows- 

\textit{~~~~~~~~~~~~~~~event(event_id, trigger_verb, actor, participant)}

\noindent
In case of a sentence from a passage, we mark each verb with an event\textunderscore id and it helps to represent the knowledge around that verb by mapping them with the event\textunderscore id. Whereas, for a question, the event\textunderscore id is unknown, so we put a variable in place of the event\textunderscore id. The trigger\textunderscore verb is in the question that triggers the predicate generation. The actor acts like the subject of the verb and the participant are the object or the modifier. The participants can be obtained from the direct object (\textit{dobj}) relations of the verb. There are various ways in which we can obtain the actor or the subject in any event.


\begin{exmp}

Given the question \textit{``What company owns Walt_Disney?''}, the three possibilities for the event predicate are given below.

1. \textit{event(E1, own, X1, O1), _similar(walt_disney, O1).}

2. \textit{event(E1, own, _, O1), _property(E1, own, _by, X1), _similar(walt_disney, O1).}

3. \textit{event(E1, own, _, _), _relation(X1, E1, _clause), _similar(walt_disney, O1).}

\noindent
Here, \textit{X1} variable binds the output and \textit{O1} variable is used internally to prove the sub-goals.

\end{exmp}

\smallskip\noindent\underline{\it Property Predicate:}
Property predicates are generated from verbs and nouns that are modified by nominal phrases in the sentence. Here, like the event predicate, we are unaware of the event_id and hence it will be set as a variable. The modified_entity is the noun or the verb that triggered the predicate generation. The preposition here can either be obtained from the case relation or can be left blank (_). The modifier is the head of the nominal modifier that can be used to constraint the query. If the modifier is the answer word, then we replace the word with the answer tag ($X_k$). 

\begin{exmp} 

Given \textit{``On what street is the ABC's headquarter located?"}, we generate: ~~~~~ \textit{_property(E2, locate, on, X2).}

\end{exmp}

\smallskip\noindent\underline{\it Similar Predicate:}
The \textit{similar} predicate models the concept of similarity between entities in which one entity is so like the other one that they can replace each other. The similar predicate thus models one of the principles of commonsense reasoning where we as humans make use of the similarity relationship while reasoning (referring to \textit{Albert Einstein} as just \textit{Einstein}). Some rules for the similar predicate are as follows:


\begin{enumerate}
    

    \item Abbreviations of concepts are similar to each other e.g., \textit{NYPD} and \textit{New York Police Dept}.
        
    \textit{~~~~~_similar(X, Y) :- _abbreviation(X, Y).}
    
    \textit{~~~~~_similar(X, Y) :- _abbreviation(Y, X).}

    \item Instances of concepts are similar to each other e.g.\textit{ Mercedes} is similar to a car

    \textit{~~~~~_similar(X, Y) :- _is(X, Y).}  

    \item Similarity follows transitivity e.g., \textit{New York}, \textit{NY}, \textit{New York City} and \textit{NYC} are all interchangeable concepts that similar to each other. 
    
    \textit{~~~~~_similar(X, Y) :- _similar(X, Z), _similar(Z, Y).}
    
\end{enumerate}

\medskip\noindent\textbf{Generating Base Constraints:}
Base constraints are generated at the end after all the constraints from the question have been generated. The term constraint here refers to constraints that arise due to the type of the answer, e.g., if the expected answer is a person, then the variable holding the answer is restricted (constrained) to one of the persons mentioned in the passage.
Constraints, thus, are sub-goals whose conjunction produces an answer to the question. Each sub-goal can be thought of as putting a constraint on the acceptable answers.
For the following cases we generate the base constraints as follows.

TIME $\rightarrow$ \textit{time($X_k$)}

DAY $\rightarrow$ \textit{day($T_k$, $X_k$), time($T_k$)}

MONTH $\rightarrow$ \textit{month($T_k$, $X_k$), time($T_k$)}

YEAR $\rightarrow$ \textit{year($T_k$, $X_k$), time($T_k$)}

PLACE $\rightarrow$ \textit{location($X_k$)} or \textit{location($X_k$, noun_location)}

PERSON $\rightarrow$ \textit{person($X_k$)} or \textit{person($X_k$, noun_person)}

\noindent In case the answer type is UNKNOWN, we may expect the answer to be a specific concept represented by the answer word. In such cases the base constraint comes from the answer word itself.

UNKNOWN $\rightarrow$ \textit{concept($X_k$)} 

e.g., company($X_k$) or city($X_k$)

\medskip\noindent\textbf{Combining constraints:} After generation of the predicates from the question and the base constraints from answer type we combine all the constraints to create the final query. This can be best explained with an example.

\begin{exmp}

For the question \textit{``When was Nikola Tesla born?''}, the following queries will be generated and automatically tried in order. 

\textit{?- event(E2, bear, S2, O2), _similar(nikola_tesla, S2),} 

\noindent \textit{~~~~~~~~~~~~property(E2, bear, on, X2), time(X2).}

\textit{?- event(E2, bear, _, O2), _property(E2, bear, _by, S2),} 

\noindent \textit{~~~~~~~~~~~~_similar(nikola_tesla, S2), property(E2, bear, on, X2), time(X2).}

\textit{?- event(E2, bear, _, _), _relation(S2, E2, _clause),}

\noindent \textit{~~~~~~~~~~~~~~~similar(nikola_tesla, S2), property(E2, bear, on, X2), time(X2).}

\textit{?- _start_date(S2, X2), _similar(nikola_tesla, S2), time(X2).}


\end{exmp}

\noindent In this sentence, we have the special predicates \_start\_date applying the constraints of time-spans on the entity  \textit{`Nikola Tesla'}.

\medskip\noindent\textbf{Query Confidence Classes:}
Ideally, we would want to be able to answer all questions with most number of constraints applied on the query so that we have a strong justification for the answer, but this is not always the case. In natural language the same questions can be posed in multiple different ways using synonymous concepts. This makes it more difficult to answer questions. Thus, it is better to make any question answering system robust enough, so that it fails gracefully. Once such attempt has been made in the query generation module by introducing the concept of confidence classes on the generated queries and by relaxing constraints on the queries to make the query more flexible. The queries that were generated in the previous section represent the most constrained queries. If an answer is found to one of those queries, then the system has the most confidence in the answer. In the absence of an answer the system starts removing constraints on the query (i.e., starts dropping sub-goals from the query) and relaxing the constraints on the query with the aim of obtaining \textit{some} answer. The queries are designed in such a way that it always produces an answer along with its confidence class. Currently, queries have been divided into 4 confidence classes that range from most confident Class I to the least confident Class IV:

\medskip 

\noindent\textbf{Certain (Class I):} These kinds of queries have all the constraints included in the natural language question. The query may contain fact predicates, subordinate constraints, answer predicates and base constraints. This class has the highest confidence level. If an answer is produced, it is a correct answer.

\noindent\textbf{Likely (Class II):} These kinds of queries are executed when a \textit{certain} answer is not produced. They do not contain any fact predicates. Fact predicates are all the predicates in the queries that do not contain any variables. Such predicates come from the question or the named entity tagger and act as constraints.

\noindent\textbf{Possible (Class III):} These kind of queries are executed when a \textit{likely} answer is not produced. They only contain the answer predicates and base constraints. This means that the fact predicates and any subordinate constraints that do not contain the answer tag ($X_k$) are removed.

\noindent\textbf{Guess (Class IV):} These kind of queries are executed when a \textit{possible} answer is not produced. They only contain the base constraints (rest of the subgoals are dropped). Answers from these queries are not very reliable as they can be thought of as guesses. This class has the lowest confidence.

\medskip\noindent 
Confidence class-based queries generated for the question \textit{``In what borough of New York City is ABC headquartered?''} are shown below:

\noindent\underline{Certain (Class I) Queries:}

\textit{?- _property(E2, borough, of, new_york_city), _property(E2, headquarter, _by, S2),} 

\textit{~~~~~~~~~~~~~~~_property(E2, headquarter, in, X2), _similar(abc, S2), borough(X2, _),} 

\textit{~~~~~~~~~~~~~~~event(E2, headquarter, _, O2), organization(abc).}

\textit{?- _property(E2, borough, of, new_york_city), _property(E2, headquarter, in, X2),}

\textit{~~~~~~~~~~~~~~~_relation(S2, E2, _clause), _similar(abc, S2), }

\textit{~~~~~~~~~~~~~~~event(E2, headquarter, _, _), organization(abc), borough(X2, _).}

\textit{?- _property(E2, borough, of, new_york_city), _property(E2, headquarter, in, X2),}

\textit{~~~~~~~~~~~~~~~_similar(abc, S2), event(E2, headquarter, S2, O2),}

\textit{~~~~~~~~~~~~~~~organization(abc), borough(X2, _).}

\noindent\underline{Likely (Class II) Queries:}

\textit{?- _property(E2, borough, of, new_york_city), _property(E2, headquarter, _by, S2),}

\textit{~~~~~~~~~~~~~~~_property(E2, headquarter, in, X2), _similar(abc, S2),}

\textit{~~~~~~~~~~~~~~~event(E2, headquarter, _, O2), borough(X2, _).}

\textit{?- _property(E2, borough, of, new_york_city), _property(E2, headquarter, in, X2),}

\textit{~~~~~~~~~~~~~~~_relation(S2, E2, _clause), _similar(abc, S2),}

\textit{~~~~~~~~~~~~~~~event(E2, headquarter, _, _), borough(X2, _).}

\textit{?- _property(E2, borough, of, new_york_city), _property(E2, headquarter, in, X2),}

\textit{~~~~~~~~~~~~~~~_similar(abc, S2), event(E2, headquarter, S2, O2), borough(X2, _).}

\noindent\underline{Possible (Class III) Queries:}

\textit{?- _property(E2, headquarter, in, X2), borough(X2, _).}

\noindent\underline{Guess (Class IV) Queries:}

\textit{?- borough(X2, _).}

These four class of queries will produce four different answers, from 100\% correct answer to a guess. Class I query response will be `Manhattan,' since ABC Corp is located in the burrough of Manhattan. That's the correct (and most specific) answer. For Class II query, the constraint `organization/1' is dropped, and so if there is another entity (not an organization) that is called ABC, then the burrough where its headquartered will be produced as the answer. For Class III query, most constraints are dropped and a burrough containing any headquarter of any entity will be named, while for Class IV query, any arbitrary burrough located anywhere will be named. We are attempting to emulate a human who tries to give a possible answer, which could be a wild guess, if he/she does not possess adequate information.

\section{CASPR: Results}
The SQuAD Dataset \cite{rajpurkar2016squad} contains more than 100,000 reading comprehensions along with questions and answers for those reading passages. SQuAD dataset uses the top 500+ articles from the English Wikipedia. These articles are then divided into paragraphs. We used the Dev Set v1.1 of the SQuAD Dataset to obtain comprehension passages for building a prototype for the proposed approach. This dataset has around 48 different articles with each article having around 50 paragraphs each. Out of the 48 different articles in the SQuAD dev set, 20 articles were chosen from different domains to help build the CASPR system (these 20 passages and associated questions can be found in \cite{github}).
These 20 articles are chosen in such a way that a lot of different domains were covered. We can roughly categorize these articles into 5 different categories: People articles, Scientific articles, Project/Event articles, Region articles and Misc. articles. Diversity is important as we have used these articles to understand the relations and that helps to craft the generic predicate rules.
Using the 20 different articles mentioned above, the ASP code was generated for one paragraph from each article. Then, ASP queries were generated for all the questions in the dataset for these paragraphs. The results show the percentage of questions for which the answer generated from the ASP solver was present in the list of answers specified for the question in the SQuAD dataset. The results are summarized in Table 2. Our results show that approximately 77.76\% of the questions are correctly answered. This shows that most of the knowledge, if not all, has been captured successfully in the ASP program generated for the passage. The ASP queries generated for the questions are very similar to the original question and convey the same meaning. 

Note that the two main reasons why a query may fail to produce an exact (certain) answer are: (i) the parser used by CASPR fails to parse the question; and, (ii) missing commonsense knowledge about concepts in the question. 
With the help of common framework, knowledge from the passage is converted into an answer set program and the ASP query is generated from the natural language question. Commonsense knowledge bridges the gap between these two (the ASP query and the clauses from passage and question respectively). In the process of automatic commonsense knowledge generation, if appropriate knowledge is not generated  for a concept present in the question, then CASPR will not be able to answer the question properly.
\begin{table}[ht]
\centering

\label{table:results1}
\begin{tabular}{|c|l|c|l|c|l|c|l|}
\hline
\multicolumn{1}{|l|}{\textbf{No}} & \textbf{Article}         & \multicolumn{1}{l|}{\textbf{Result}} & \textbf{\%} & \multicolumn{1}{l|}{\textbf{No}} & \textbf{Article}             & \multicolumn{1}{l|}{\textbf{Result}} & \textbf{\%}             \\ \hline
1                                 & ABC Corp.                & 3/4                                  & 100         & 11                               & Kenya                        & 5/5                                  & 100                     \\ \hline
2                                 & Amazon Rainforest        & 12/14                                & 85.7        & 12                               & Martin Luther                & 2/5                                  & 40                      \\ \hline
3                                 & Apollo                   & 4/5                                  & 80          & 13                               & Nikola Tesla                 & 6/7                                  & 85.7                    \\ \hline
4                                 & Chloroplasts             & 4/5                                  & 80          & 14                               & Normans                      & 4/5                                  & 80                      \\ \hline
5                                 & Computational Complexity & 3/3                                  & 100         & 15                               & Oxygen                       & 8/15                                 & 53.3                    \\ \hline
6                                 & Ctenophora               & 9/12                                 & 75          & 16                               & Rhine                        & 5/8                                  & 62.5                    \\ \hline
7                                 & European Union Law       & 13/13                                & 100         & 17                               & Southern California          & 3/5                                  & 60                      \\ \hline
8                                 & Genghis Khan             & 3/5                                  & 60          & 18                               & Steam Engine                 & 4/5                                  & 80                      \\ \hline
9                                 & Geology                  & 4/5                                  & 80          & 19                               & Super Bowl 50                & 25/29                                & 86.2                    \\ \hline
10                                & Immune System            & 13/15                                & 86.6        & 20                               & Warsaw                       & 3/5                                  & \multicolumn{1}{c|}{60} \\ \hline
\multicolumn{5}{|l|}{\textbf{Total}}                                                                                                                 & \multicolumn{1}{c|}{135/171} & \multicolumn{2}{c|}{78.95\%}                                   \\ \hline
\multicolumn{5}{|l|}{\textbf{Average Result}}                                                                                                        & \multicolumn{3}{c|}{77.76\%}                                                                  \\ \hline
\end{tabular}
\caption{Results for Question answering}
\end{table}

Note that our system has shown excellent execution performance on the 171 questions on 20 passages tried thus far: 80\% of the questions were answered in 2 to 3 milliseconds, while the rest were answered in a few seconds.

\section{Contributions}
The main contribution of this paper is an effective and efficient method for converting textual data into knowledge represented as an answer set program that can be processed on our query-driven s(ASP) ASP system. 
This includes developing a neo-davidsonian logic inspired generic calculus that helps represent knowledge, and using knowledge sources such as WordNet to acquire commonsense knowledge about terms found in the text to create a custom ontology for the problem. 
This automatic custom ontology generation process makes the CASPR system scalable.
Yet another novelty is in showing how word sense disambiguation can be elegantly modeled using ASP. Our system is based on  `\textit{truly understanding}' the text just as humans do. 

The paper also proposes a framework for converting natural language questions into ASP queries. These queries can be run on the query-driven s(ASP) system to compute answers. The query generation framework is made robust through broadening of queries by dropping constraints, thus increasing the possibility that the question will indeed be answered (even though in the worst case the answer may just be a guess). This approach to handling question answering is yet another novelty of the proposed system.

\section{Related Works}
With respective to related work, Cyc \cite{wiki:xxx} is one of the oldest AI project that attempts to model commonsense reasoning. In Cyc, knowledge is presented in the form of a vast collection of ontologies that consist of implicit knowledge and rules about the world that represents commonsense knowledge. 
The ontology of Cyc as of 2017 contains around 1,500,000 terms. This includes around 500,000 collections, 50,000+ predicates and around a million well known entities. Cyc's language CycL made it efficient to represent commonsense knowledge in the project and determined how this knowledge is represented in the project. 
Cyc uses a \textit{community of agents} consisting of multiple reasoning agents that rely on more than 1000 heuristic modules to solve inference problems. Cyc, however, does not work with natural language, though recently some efforts have been started. A potential problem with Cyc is knowing which agent to apply, and which heuristic to use. For a commonsense reasoning system to be successful, \textit{it has to be modeled in a very simple way}. Otherwise, we may possess the individual pieces of knowledge to answer a given question, but may not be able to compose these pieces together to arrive at an answer. For this reason, CASPR represents knowledge using very few generic predicates and uses simple ASP-based reasoning patterns to compute answers. Our initial experiments suggest that our approach is effective.   

The application of ASP to solve NLP tasks is not new. Vo and Baral \cite{Vo-Baral} have developed the NL2KR tool that allows natural language text to be translated to an answer set program. Similarly, the work by Costantini and Paolucci \cite{costantini} shows how ASP code can be generated automatically from natural language using lambda calculus. Action language based approaches are also popular in translating natural language to ASP \cite{text2alm,restkb} etc. Several other researchers have worked on applying ASP for NLP tasks and their efforts are reported in the first workshop on NLP and Automated Reasoning \cite{aspnlp}. Our approach has elements common with these efforts, however, our work is based on the query-driven s(ASP) predicate ASP engine, and thus is scalable and not constrained by limitations of grounding based implementations of ASP. The query-driven s(ASP) system is crucial to our success as, like the current work, our previous works in this direction show promising results in the domain of visual question answering \cite{aqua} and natural language question answering \cite{square,basu2021aaai}.

There are many approaches to question answering based on machine learning (cf. SQuAD website \cite{rajpurkar2016squad}). However, they are not based on \textit{actually understanding} the text and so can only answer questions related to data they are trained on. CASPR, in contrast, \textit{understands} the knowledge contained in the text and has shown promising results for a subset of the SQuAD dataset. 
It should be noted that machine learning based approaches are based on statistical techniques and pattern matching, and do not represent the knowledge inherent in the text in any form. There is nothing that corresponds to \textit{understanding} of the text, in contrast to our approach where this \textit{understanding} is represented as knowledge modeled in ASP. For this reason, we do not elaborate too much here on machine learning based approaches to question answering. 
Also, there are some machine comprehension efforts based on semantic lexicon \cite{verbnetto} that work really well on procedural texts, however they may not work well on non-procedural, descriptive text. In addition,  they do not perform non-monotonic reasoning. In contrast, CASPR can handle both non-procedural texts and perform non-monotonic reasoning.



A critical component of CASPR's success is the s(ASP) query-driven, predicate ASP system that leads to 
three major advantages for CASPR: (i) only parts of the knowledge base relevant to answering the question are explored during execution; (ii) justification for answers can be extracted from the justification tree produced by s(ASP); and, (iii) the question answering system is scalable, as no grounding of the program needs to be done as s(ASP) can execute predicates directly under the stable model semantics.

\section{Conclusions and Future Work}

In this paper we presented a comprehensive ASP-based  framework called CASPR for natural language question answering. CASPR translates natural language text and questions into ASP code and ASP query, respectively. It also imports commonsense knowledge from other sources such as WordNet. We also presented an elegant solution for word sense disambiguation based on default and classical negation. We used the query driven s(ASP) engine to perform commonsense reasoning to obtain the answer. 
CASPR was applied to the SQuAD dataset and has shown promising results. Our approach always finds the most suitable answer, if adequate knowledge is present and the natural language analysis works correctly. If not, it employs a simple method for weakening the question to ensure that an answer will be found, even though it may be a guess. CASPR not only  gives promising results, it also enjoys other benefits such as explainability, generalizability, and interpretability.

CASPR is a step toward building truly intelligent systems that mimic human reasoning based on ``truly understanding" the text.
Our future work includes (i) extending the system to handle more complex questions (e.g., causality questions), (ii) incorporating additional knowledge resources (e.g., Wikidata \cite{wikidata} and Dbpedia \cite{dbpedia}) for importing more commonsense knowledge to handle a broader class of questions, (iii) adding more semantics in knowledge representations, such as \textit{semantic relations} or pre-defined semantic predicates (e.g., VerbNet)  (iv) generating justifications to a question's answer in a more human-readable way, (v) extending the system for other NLP tasks: text summarization, question generation, etc.

\medskip

\noindent{\bf Acknowledgement:}
Thanks to Sarat Varanasi, Elmer Salazar, Joaquin Arias, Zhuo Chen, and Serdar Erbatur for discussions and help. Authors gratefully acknowledge support from NSF grants IIS 1910131 and IIS 1718945.
\smallskip

\bibliographystyle{acmtrans}
\bibliography{bibliography}

\begin{thebibliography}{}

\bibitem[\protect\citeauthoryear{Alviano, Faber, Leone, Perri, Pfeifer, and
  Terracina}{Alviano et~al\mbox{.}}{2011}]{alviano2011disjunctive}
{\sc Alviano, M.}, {\sc Faber, W.}, {\sc Leone, N.}, {\sc Perri, S.}, {\sc
  Pfeifer, G.}, {\sc and} {\sc Terracina, G.} 2011.
\newblock The disjunctive datalog system dlv.
\newblock In {\em Datalog Reloaded}. Springer, 282--301.

\bibitem[\protect\citeauthoryear{Arias, Carro, Salazar, Marple, and
  Gupta}{Arias et~al\mbox{.}}{2018}]{arias2018constraint}
{\sc Arias, J.}, {\sc Carro, M.}, {\sc Salazar, E.}, {\sc Marple, K.}, {\sc
  and} {\sc Gupta, G.} 2018.
\newblock Constraint answer set programming without grounding.
\newblock {\em arXiv preprint arXiv:1804.11162\/}.

\bibitem[\protect\citeauthoryear{Auer, Bizer, Kobilarov, Lehmann, Cyganiak, and
  Ives}{Auer et~al\mbox{.}}{2007}]{dbpedia}
{\sc Auer, S.}, {\sc Bizer, C.}, {\sc Kobilarov, G.}, {\sc Lehmann, J.}, {\sc
  Cyganiak, R.}, {\sc and} {\sc Ives, Z.} 2007.
\newblock Dbpedia: A nucleus for a web of open data.
\newblock In {\em The semantic web}. Springer, 722--735.

\bibitem[\protect\citeauthoryear{Baral}{Baral}{2003}]{baral2003knowledge}
{\sc Baral, C.} 2003.
\newblock {\em Knowledge representation, reasoning and declarative problem
  solving}.
\newblock Cambridge university press.

\bibitem[\protect\citeauthoryear{Baral and Sch{\"{u}}ller}{Baral and
  Sch{\"{u}}ller}{2013}]{aspnlp}
{\sc Baral, C.} {\sc and} {\sc Sch{\"{u}}ller, P.}, Eds. 2013.
\newblock {\em Proceedings of the 1st Workshop on Natural Language Processing
  and Automated Reasoning 2013}. {CEUR} Workshop Proceedings, vol. 1044.
  CEUR-WS.org.

\bibitem[\protect\citeauthoryear{Basu, Shakerin, and Gupta}{Basu
  et~al\mbox{.}}{2020}]{aqua}
{\sc Basu, K.}, {\sc Shakerin, F.}, {\sc and} {\sc Gupta, G.} 2020.
\newblock Aqua: Asp-based visual question answering.
\newblock In {\em International Symposium on Practical Aspects of Declarative
  Languages}. Springer, 57--72.

\bibitem[\protect\citeauthoryear{Basu, Varanasi, Shakerin, Arias, and
  Gupta}{Basu et~al\mbox{.}}{2021}]{basu2021aaai}
{\sc Basu, K.}, {\sc Varanasi, S.}, {\sc Shakerin, F.}, {\sc Arias, J.}, {\sc
  and} {\sc Gupta, G.} 2021.
\newblock Knowledge-driven natural language understanding of english text and
  its applications.
\newblock {\em arXiv preprint arXiv:2101.11707\/}.

\bibitem[\protect\citeauthoryear{Basu, Varanasi, Shakerin, and Gupta}{Basu
  et~al\mbox{.}}{2020}]{square}
{\sc Basu, K.}, {\sc Varanasi, S.~C.}, {\sc Shakerin, F.}, {\sc and} {\sc
  Gupta, G.} 2020.
\newblock Square: Semantics-based question answering and reasoning engine.
\newblock {\em arXiv preprint arXiv:2009.10239\/}.

\bibitem[\protect\citeauthoryear{Chen, Marple, Salazar, Gupta, and Tamil}{Chen
  et~al\mbox{.}}{2016}]{zhuo-iclp2016}
{\sc Chen, Z.}, {\sc Marple, K.}, {\sc Salazar, E.}, {\sc Gupta, G.}, {\sc and}
  {\sc Tamil, L.} 2016.
\newblock A physician advisory system for chronic heart failure management
  based on knowledge patterns.
\newblock {\em Theory and Practice of Logic Programming\/}~{\em 16,\/}~5-6,
  604--618.

\bibitem[\protect\citeauthoryear{Clark, Dalvi, and Tandon}{Clark
  et~al\mbox{.}}{2018}]{verbnetto}
{\sc Clark, P.}, {\sc Dalvi, B.}, {\sc and} {\sc Tandon, N.} 2018.
\newblock What happened? leveraging verbnet to predict the effects of actions
  in procedural text.
\newblock {\em arXiv preprint arXiv:1804.05435\/}.

\bibitem[\protect\citeauthoryear{Costantini and Paolucci}{Costantini and
  Paolucci}{2010}]{costantini}
{\sc Costantini, S.} {\sc and} {\sc Paolucci, A.} 2010.
\newblock Towards translating natural language sentences into asp.
\newblock In {\em CILC}.

\bibitem[\protect\citeauthoryear{Davidson}{Davidson}{1984}]{neodavidsonian}
{\sc Davidson, D.} 1984.
\newblock {\em Inquiries into Truth and Interpretation}.
\newblock Oxford University Press.

\bibitem[\protect\citeauthoryear{De~Marneffe, Dozat, Silveira, Haverinen,
  Ginter, Nivre, and Manning}{De~Marneffe
  et~al\mbox{.}}{2014}]{typeddependencies1}
{\sc De~Marneffe, M.-C.}, {\sc Dozat, T.}, {\sc Silveira, N.}, {\sc Haverinen,
  K.}, {\sc Ginter, F.}, {\sc Nivre, J.}, {\sc and} {\sc Manning, C.~D.} 2014.
\newblock Universal stanford dependencies: A cross-linguistic typology.
\newblock In {\em LREC}. Vol.~14. 4585--4592.

\bibitem[\protect\citeauthoryear{De~Marneffe and Manning}{De~Marneffe and
  Manning}{2008}]{typeddependencies}
{\sc De~Marneffe, M.-C.} {\sc and} {\sc Manning, C.~D.} 2008.
\newblock Stanford typed dependencies manual.
\newblock Tech. rep., Technical report, Stanford University.

\bibitem[\protect\citeauthoryear{Finkel, Grenager, and Manning}{Finkel
  et~al\mbox{.}}{2005}]{nertagger}
{\sc Finkel, J.~R.}, {\sc Grenager, T.}, {\sc and} {\sc Manning, C.} 2005.
\newblock Incorporating non-local information into information extraction
  systems by gibbs sampling.
\newblock In {\em Proceedings of the 43rd annual meeting on association for
  computational linguistics}. Association for Computational Linguistics,
  363--370.

\bibitem[\protect\citeauthoryear{Gebser, Kaminski, Kaufmann, Ostrowski, Schaub,
  and Thiele}{Gebser et~al\mbox{.}}{2010}]{gebser2010gringo}
{\sc Gebser, M.}, {\sc Kaminski, R.}, {\sc Kaufmann, B.}, {\sc Ostrowski, M.},
  {\sc Schaub, T.}, {\sc and} {\sc Thiele, S.} 2010.
\newblock gringo, clasp, clingo, and iclingo.
\newblock {\em user guide\/}.

\bibitem[\protect\citeauthoryear{Gelfond and Kahl}{Gelfond and
  Kahl}{2014}]{gelfond2014knowledge}
{\sc Gelfond, M.} {\sc and} {\sc Kahl, Y.} 2014.
\newblock {\em Knowledge representation, reasoning, and the design of
  intelligent agents: The answer-set programming approach}.
\newblock Cambridge University Press.

\bibitem[\protect\citeauthoryear{Gelfond and Lifschitz}{Gelfond and
  Lifschitz}{1988}]{gelfond1988stable}
{\sc Gelfond, M.} {\sc and} {\sc Lifschitz, V.} 1988.
\newblock The stable model semantics for logic programming.
\newblock In {\em ICLP/SLP}. Vol.~88. 1070--1080.

\bibitem[\protect\citeauthoryear{Jonson-Laird}{Jonson-Laird}{2009}]{how-we-reason}
{\sc Jonson-Laird, P.} 2009.
\newblock {\em How We Reason}.
\newblock Oxford University Press.

\bibitem[\protect\citeauthoryear{Jurafsky}{Jurafsky}{2000}]{nlpbook}
{\sc Jurafsky, D.} 2000.
\newblock {\em Speech \& language processing}.
\newblock Pearson Education India.

\bibitem[\protect\citeauthoryear{Kipper, Korhonen, Ryant, and Palmer}{Kipper
  et~al\mbox{.}}{2006}]{verbnet}
{\sc Kipper, K.}, {\sc Korhonen, A.}, {\sc Ryant, N.}, {\sc and} {\sc Palmer,
  M.} 2006.
\newblock Extending verbnet with novel verb classes.
\newblock In {\em Proceedings of the Fifth International Conference on Language
  Resources and Evaluation, {LREC} 2006, Genoa, Italy, May 22-28, 2006.}
  1027--1032.

\bibitem[\protect\citeauthoryear{Mahdisoltani, Biega, and
  Suchanek}{Mahdisoltani et~al\mbox{.}}{2015}]{yago}
{\sc Mahdisoltani, F.}, {\sc Biega, J.}, {\sc and} {\sc Suchanek, F.~M.} 2015.
\newblock {YAGO3:} {A} knowledge base from multilingual wikipedias.
\newblock In {\em {CIDR} 2015, Seventh Biennial Conference on Innovative Data
  Systems Research, Asilomar, CA, USA, January 4-7, 2015, Online Proceedings}.

\bibitem[\protect\citeauthoryear{Manning, Surdeanu, Bauer, Finkel, Bethard, and
  McClosky}{Manning et~al\mbox{.}}{2014}]{manning-EtAl:2014:P14-5}
{\sc Manning, C.~D.}, {\sc Surdeanu, M.}, {\sc Bauer, J.}, {\sc Finkel, J.},
  {\sc Bethard, S.~J.}, {\sc and} {\sc McClosky, D.} 2014.
\newblock The {Stanford} {CoreNLP} natural language processing toolkit.
\newblock In {\em ACL System Demonstrations}. 55--60.

\bibitem[\protect\citeauthoryear{Marple, Salazar, and Gupta}{Marple
  et~al\mbox{.}}{2017}]{marple2017computing}
{\sc Marple, K.}, {\sc Salazar, E.}, {\sc and} {\sc Gupta, G.} 2017.
\newblock Computing stable models of normal logic programs without grounding.
\newblock {\em arXiv preprint arXiv:1709.00501\/}.

\bibitem[\protect\citeauthoryear{Miller}{Miller}{1995}]{miller1995wordnet}
{\sc Miller, G.~A.} 1995.
\newblock {WordNet}: a lexical database for english.
\newblock {\em Communications of the ACM\/}~{\em 38,\/}~11, 39--41.

\bibitem[\protect\citeauthoryear{Mitchell, Cohen, Hruschka, Talukdar,
  Betteridge, Carlson, Dalvi, Gardner, Kisiel, Krishnamurthy, Lao, Mazaitis,
  Mohamed, Nakashole, Platanios, Ritter, Samadi, Settles, Wang, Wijaya, Gupta,
  Chen, Saparov, Greaves, and Welling}{Mitchell et~al\mbox{.}}{2015}]{nell}
{\sc Mitchell, T.}, {\sc Cohen, W.}, {\sc Hruschka, E.}, {\sc Talukdar, P.},
  {\sc Betteridge, J.}, {\sc Carlson, A.}, {\sc Dalvi, B.}, {\sc Gardner, M.},
  {\sc Kisiel, B.}, {\sc Krishnamurthy, J.}, {\sc Lao, N.}, {\sc Mazaitis, K.},
  {\sc Mohamed, T.}, {\sc Nakashole, N.}, {\sc Platanios, E.}, {\sc Ritter,
  A.}, {\sc Samadi, M.}, {\sc Settles, B.}, {\sc Wang, R.}, {\sc Wijaya, D.},
  {\sc Gupta, A.}, {\sc Chen, X.}, {\sc Saparov, A.}, {\sc Greaves, M.}, {\sc
  and} {\sc Welling, J.} 2015.
\newblock Never-ending learning.
\newblock In {\em Proceedings of the Twenty-Ninth AAAI Conference on Artificial
  Intelligence (AAAI-15)}.

\bibitem[\protect\citeauthoryear{Olson and Lierler}{Olson and
  Lierler}{2019}]{text2alm}
{\sc Olson, C.} {\sc and} {\sc Lierler, Y.} 2019.
\newblock Information extraction tool text2alm: From narratives to action
  language system descriptions.
\newblock {\em Electronic Proceedings in Theoretical Computer Science\/}~{\em
  306}, 87–100.

\bibitem[\protect\citeauthoryear{Pendharkar}{Pendharkar}{2018a}]{dhruvathesis}
{\sc Pendharkar, D.} 2018a.
\newblock {\em An Answer Set Programming based Approach to Representing and
  Querying Textual Knowledge}.
\newblock M.S. Thesis, The University of Texas at Dallas,
  \url{http://utdallas.edu/~gupta/dpthesis.pdf}.

\bibitem[\protect\citeauthoryear{Pendharkar}{Pendharkar}{2018b}]{github}
{\sc Pendharkar, D.} 2018b.
\newblock Caspr.
\newblock \url{https://github.com/DhruvaPendharkar/thesis-project}.

\bibitem[\protect\citeauthoryear{Rajpurkar, Zhang, Lopyrev, and
  Liang}{Rajpurkar et~al\mbox{.}}{2016}]{rajpurkar2016squad}
{\sc Rajpurkar, P.}, {\sc Zhang, J.}, {\sc Lopyrev, K.}, {\sc and} {\sc Liang,
  P.} 2016.
\newblock {SQuAD}: 100,000+ questions for machine comprehension of text.
\newblock {\em arXiv preprint arXiv:1606.05250\/}.

\bibitem[\protect\citeauthoryear{Schuster and Manning}{Schuster and
  Manning}{2016}]{schuster2016enhanced}
{\sc Schuster, S.} {\sc and} {\sc Manning, C.~D.} 2016.
\newblock Enhanced english universal dependencies: An improved representation
  for natural language understanding tasks.
\newblock In {\em LRED'16}. 2371--2378.

\bibitem[\protect\citeauthoryear{Vo, Mitra, and Baral}{Vo
  et~al\mbox{.}}{2015}]{Vo-Baral}
{\sc Vo, N.~H.}, {\sc Mitra, A.}, {\sc and} {\sc Baral, C.} 2015.
\newblock The {NL2KR} platform for building natural language translation
  systems.
\newblock In {\em Proceedings of the 53rd Annual Meeting of the Association for
  Computational Linguistics and the 7th International Joint Conference on
  Natural Language Processing of the Asian Federation of Natural Language
  Processing, {ACL} 2015, July 26-31, 2015, Beijing, China, Volume 1: Long
  Papers}. 899--908.

\bibitem[\protect\citeauthoryear{Vrande{\v{c}}i{\'c} and
  Kr{\"o}tzsch}{Vrande{\v{c}}i{\'c} and Kr{\"o}tzsch}{2014}]{wikidata}
{\sc Vrande{\v{c}}i{\'c}, D.} {\sc and} {\sc Kr{\"o}tzsch, M.} 2014.
\newblock Wikidata: a free collaborative knowledgebase.
\newblock {\em Communications of the ACM\/}~{\em 57,\/}~10, 78--85.

\bibitem[\protect\citeauthoryear{{Wikipedia contributors}}{{Wikipedia
  contributors}}{2018}]{wiki:xxx}
{\sc {Wikipedia contributors}}. 2018.
\newblock Cyc --- {Wikipedia}{,} the free encyclopedia.
\newblock [Online; accessed 17-May-2018].

\bibitem[\protect\citeauthoryear{Zhang, Benton, and Inclezan}{Zhang
  et~al\mbox{.}}{2020}]{restkb}
{\sc Zhang, Q.}, {\sc Benton, C.}, {\sc and} {\sc Inclezan, D.} 2020.
\newblock An application of asp theories of intentions to understanding
  restaurant scenarios: Insights and narrative corpus.
\newblock {\em Theory and Practice of Logic Programming\/}~{\em 20,\/}~2,
  273--293.

\end{thebibliography}

\label{lastpage}
\end{document}